\documentclass[sigconf, nonacm]{acmart}

\usepackage{amsmath}
\usepackage{graphicx}

\usepackage{placeins}
\AtBeginDocument{%
  \providecommand\BibTeX{{%
    \normalfont B\kern-0.5em{\scshape i\kern-0.25em b}\kern-0.8em\TeX}}}

\begin{document}

\title{GenAI Assisting Medical Training}

\author{Stefan Gerd Fritsch, Matthias Tschöpe, Vitor Fortes Rey}
\affiliation{%
  \institution{DFKI}
  \streetaddress{Trippstadter Straße 122}
  \city{Kaiserslautern}
  \country{Germany}
  \postcode{67663} }

  \author{Lars Krupp, Agnes Grünerbl}
\affiliation{%
  \institution{RPTU and DFKI}
  \streetaddress{Trippstadter Straße 122}
  \city{Kaiserslautern}
  \country{Germany}
  \postcode{67663} }

  \author{Eloise Monger, Sarah Trevenna}
\affiliation{%
  \institution{University of Southampton}
  \city{Southampton}
  \country{UK}
  }
\renewcommand{\shortauthors}{DFKI et al.}

\begin{abstract}
Medical procedures such as venipuncture and cannulation are essential for nurses and require precise skills. Learning this skill, in turn, is a challenge for educators due to the number of teachers per class and the complexity of the task. The study aims to help students with skill acquisition and alleviate the educator's workload by integrating generative AI methods to provide real-time feedback on medical procedures such as venipuncture and cannulation.
\end{abstract}



\maketitle

\section{Introduction}
 Performing venipunctures (the procedure of taking a blood sample from a patient) \cite{venipuncture22} and cannulations (the procedure of inserting a cannula for infusions and intravenous medication) \cite{cannulation18} is an essential part of a nurse's work. As this is an invasive and non-trivial act, it is essential for nurses training to acquire this skill. In the nurse training program, students are trained using a simulation training arm. These tasks require fine motor skills and involve multiple steps to prevent infection and reduce pain, bruising, and complications. Providing feedback for these procedures is challenging as expert guidance can be variable and there is often a lack of accuracy in the complex steps and precision in the angle, smoothness, and speed of insertion.
 The main challenge in training such medical procedures, despite the availability of training arms, is the teachers lack the time for 101 training. Most of the time such training classes comprise 15-20 students with one teacher, who often is busy helping one or two students, while the rest try to figure it out themselves. This is not optimal for such an invasive medical procedure.
 
\subsection{Idea and Objectives}
In several discussions with nurse trainers, the idea arose to combine nurse training with generative AI methods to build a live feedback system for nurse students. Using data from training videos, sensor recordings, and an appropriate language model (LLM), our goal is to offer personalized feedback to nursing trainees and improve their learning experience.
While the existing model operates by analyzing recorded sessions, the ambition is to transition to live feedback. This would mean that trainees and instructors receive insights and guidance live, during actual training sessions.
\begin{figure}
    \centering
    \includegraphics[width=0.4\textwidth]{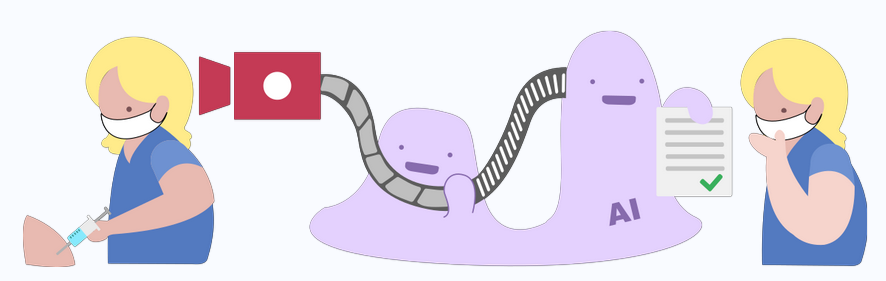}
    \caption{AI providing feedback on performed procedure}
    \label{fig:enter-label}
    \vspace{-3mm}
\end{figure}

\section{Dataset} 
During a three-day period from 27 to 29 November 2023, the data collection took place at the University of Southampton. The focus was on two procedures: Cannulation and Venipuncture, with a total of 20 people taking part. This included 13 students with varying levels of experience and 7 experts. Each participant performed both procedures twice, resulting in a total of 80 recordings.

The collected dataset includes several components:
\begin{itemize}
     \item Static cameras: three static cameras (high-resolution video (1920x1080) at 60 Hz), positioned at strategic locations around the bed (foot, head, and side).
     \item GoPro camera: capture user view.
     \item Audio recordings: to ensure synchronized audio-visual data.
     \item IMU data: Two Apple smartwatches worn by participants on their left and right wrists provided IMU (Inertial Measurement Unit) data, which was used to capture the nuances of hand movements during the procedures.
     \item Feedback forms:  An expert observer filled in professional feedback forms about the performance of the study participants while participants performed the procedures. Participants filled in a form regarding their demographics
\end{itemize} 

\section{Data Preprocessing}
\subsection{Synchronization:} To ensure the synchronization of all cameras, including three static cameras and a GoPro, we had each participant clap three times before starting their procedure. The sound peaks of the clapping were used as reference points for synchronizing the video footage. The data recorded with the smartwatch will be synchronized and added at a later step.

\subsection{Labeling:} Following the official cannulation and venipuncture protocols, we developed detailed labeling instructions. These instructions outline specific, easily identifiable points that mark the beginning and end of each action. We then labeled the videos as accurately as possible (on a frame level), i.e., the start and end times of each action/procedure step. For example:
        \begin{itemize}
            \item \textbf{Step:} Apply tourniquet to arm
            \item \textbf{Start Action:} Pick up tourniquet \textbar\textbar\ touch tourniquet
            \item \textbf{End Action:} Stop touching tourniquet
        \end{itemize}

\subsection{Transcription:} The audio feedback provided by the observers was transcribed into text similar to \cite{radford2023robust}, using Nuance Dragon 15 Professional for the automatic transcription, but was manually corrected afterwards if needed. This transcription is important as it is later used as input for the AI model or Large Language Model (LLM) to generate feedback for the participants.


\section{First Steps}
\subsection{Video Classification}
 In the first step, we are currently focusing on the development of methods for video classification, in particular the recognition of the steps performed in the videos concerning the specifications from the medical protocols. This involves the precise identification of the individual actions performed during the procedures.

\subsection{Developing a Method for Providing Feedback}
 In the first steps, methods for providing feedback will also be developed. One approach is to use a Large Language Model (LLM), which provides contextual information about how the procedure should be performed together with the results of the video classification process, to generate feedback for the user.
The first method, which was tested with videos (video only, later smartwatch data should also be included), comprises several steps:
\begin{enumerate}
     \item  \textbf{ Splitting the dataset: } We split the dataset into training, validation, and test sets at the subject level to ensure subject independence.
    \item  \textbf{Data sampling:} We apply an overlapping sliding window technique to the videos to extract data samples for training the models.
     \item \textbf{Video Classification:}  We are working on fine-tuning video classification models, such as S3D \cite{xie2018rethinking}, to predict the labels corresponding to the steps of the protocol.
    \item  \textbf{LLM integration: } A medical LLM should be integrated that is stimulated with procedural context information and feedback from observers as examples, as well as the results of video classification. This will allow the model to provide feedback on whether the procedure has been performed correctly, including checking the order of steps, identifying missing steps, and ensuring sufficient waiting time after cleaning the skin.
\end{enumerate}
By integrating these components, we hope to create a robust system capable of providing accurate and helpful feedback to users based on their performance of medical procedures.


 \begin{figure}
    \centering
    \includegraphics[width=0.4\textwidth]{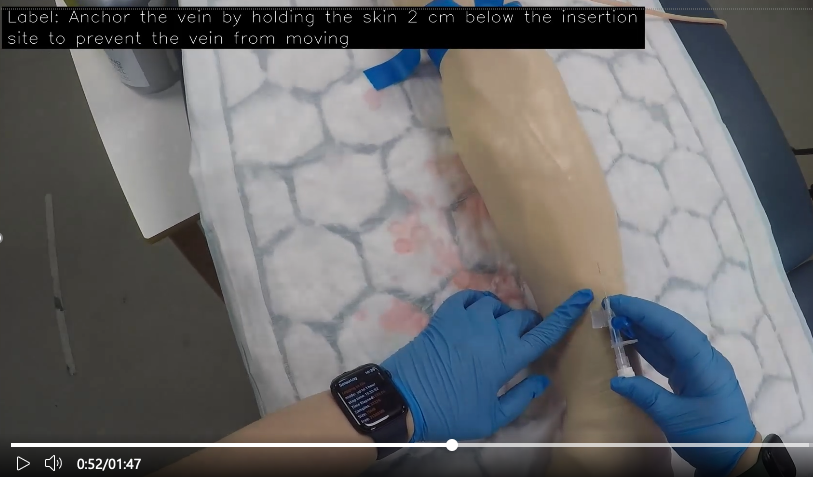}
    \caption{Labeling each step in the videos}
    \label{fig:enter-label}
\end{figure}

\section{Outlook and Next Steps:}
In the next steps, we plan to improve our system by incorporating smartwatch data as an additional input source. The aim is for the system to not only recognize the sequence of steps but also the nuances in the execution of each step, such as the angle of the needle insertion. We currently lack the means to accurately measure these details. However, we have the necessary data, including feedback from observers and forms, with which we can train models that can assess the quality of each step.

In addition, we intend to evaluate our method in a realistic, real-world scenario through a follow-up study at the University of Southampton by the end of this year. 

\section{Discussion and Conclusion}
Several important conclusions have emerged from our study so far. Correct labeling of medical procedures in videos is a challenging task, even for humans, as understanding medical procedures is necessary. This raises the question of whether and to what extent Generative AI could help.

\section{Acknowledgments}   
We want to acknowledge that ChatGPT-3.5 has been used to restructure some sentences and in the initial brainstorming following current ACM policies. This work is partially funded by the Humane AI Net Project (EU H2020-ICT-48 Grant No. 952026)
\bibliographystyle{ACM-Reference-Format}
\bibliography{sample}

\end{document}